# Bridging the Clinical Expertise Gap: Development of a Web-Based Platform for Accessible Time Series Forecasting and Analysis

Aaron D. Mullen, M.S.[1], Daniel R. Harris, Ph. D.[2], Svetla Slavova, Ph. D.[3], V.K. Cody Bumgardner, Ph.D.[1]
[1]Center for Applied AI, University of Kentucky, Lexington, Kentucky, USA
[2]College of Medicine, University of Kentucky, Lexington, Kentucky, USA
[3]Department of Biostatistics, University of Kentucky, Lexington, Kentucky, USA

**Abstract**

*Time series forecasting has applications across domains and industries, especially in healthcare, but the technical expertise required to analyze data, build models, and interpret results can be a barrier to using these techniques. This article presents a web platform that makes the process of analyzing and plotting data, training forecasting models, and interpreting and viewing results accessible to researchers and clinicians. Users can upload data and generate plots to showcase their variables and the relationships between them. The platform supports multiple forecasting models and training techniques which are highly customizable according to the user's needs. Additionally, recommendations and explanations can be generated from a large language model that can help the user choose appropriate parameters for their data and understand the results for each model. The goal is to integrate this platform into learning health systems for continuous data collection and inference from clinical pipelines.*

**Introduction**

Time series forecasting encompasses both statistical and machine learning methods for a wide variety of uses across different domains. Forecasting techniques can be used for predicting stock prices and market trends[1,2], planning power supply in different areas based on demand forecasts[3], predicting the weather to make operational decisions about renewable energy systems[4], predicting flooding and managing water resources[5], and understanding how external factors influence retail sales[6]. Time series forecasting is especially prevalent in healthcare at both a population level, such as for epidemiology and predicting pandemic-related data[7], and at an individual level, like using prediction techniques for personalized medicine based on continuous monitoring[8].

Technical backgrounds are typically required to design and build the models to generate predictions and to interpret the results of those predictions. Forecasting is primarily done with either statistical or machine learning methods. For statistical models, such as Autoregressive Integrated Moving Average (ARIMA), a background in statistics is required to understand the requirements of data distribution and feature selection. For machine learning models, such as Long Short-Term Memory neural networks[9], a background in computer science with a specialization in machine learning is typically required to understand model architectures and parameter tuning. While machine learning models typically offer improved accuracy and predictive power when compared to statistical techniques, this can come at the cost of interpretability[10,11]. Neither type of model is perfect on its own, and implementing these models effectively requires deep understanding of the data and methods being used, frequently making it difficult to optimally utilize these techniques[12].

There are a variety of different tools and methods accessible to researchers and clinicians to perform time series forecasting, but these tools have their own limitations. Some are incredibly powerful and customizable, such as Python libraries like statsmodels[13], scikit-learn[14], and PyTorch[15]. However, these libraries are only usable by individuals with programming expertise, and even then, they can be difficult to correctly implement. On the other hand, while tools such as Excel are very accessible and simple to use, they are limited in their capabilities to perform detailed custom analysis. Thus, a tool for time series forecasting that is both powerful and accessible is needed in this field.

There have been other attempts to address this gap in the domain of time series forecasting. PyCaret is an example of a "low-code" library that allows users to implement models easier with only a few lines of code, but this still necessitates some understanding of programming workflows[16]. There are also commercial alternatives that can handle time series forecasting through web-based interfaces, such as Amazon Forecast[17], DataRobot[18], or Google Cloud[19], but these tools are proprietary and often cost-prohibitive. Combining large language models (LLMs) with time series forecasting is a cutting-edge research topic with some existing work, but this is primarily focused on using the LLM to generate forecasts themselves[20,21], which differs from the use case described in this article.

This article presents the Forecaster, an open-source, web-based platform that incorporates data pre-processing, model training, and results interpretation into a single, easy-to-use interface. Through this platform, users can upload a dataset and easily generate line plots or Gantt charts for any of their variables, or multiple variables at once, to identify trends and compare columns. After this, users can select which models they want to train and can modify parameters for those models. To better assist the user with making these selections, an LLM assistant is provided that can use summary statistics about the uploaded dataset to make recommendations on parameter choices and models, using stats such as dataset size, time periods, and distributions. Once the user makes these choices, they can begin the training job, which will use those parameters to train and evaluate each of the models. Once this job is finished, users can view results on the site, including performance metrics for each model and graphs showing what the predictions look like. Again, an LLM is incorporated to provide explanations for the results and make recommendations about which models would be best to use. The user can also choose to easily re-run any training jobs with modified parameters or generate new forecasts for the future with previously trained models. Overall, this system handles many different aspects of the process of generating predictions in an automated, user-friendly way, making these powerful tools more accessible to clinicians and researchers.

An additional goal with the development of the Forecaster is the integration of this platform with continuous health monitoring systems. Trained models can be part of a larger learning health system, where data is passively collected from clinical pipelines and used to continually update forecasting models. For example, eICU data streams could be fed into the system via API for repeated re-training and evaluation, ensuring continuous monitoring and updated forecasting using vitals from patients.

**Methods**

The Forecaster is built on a modular, dual-backend architecture. The primary frontend web-serving processes, such as user management and database interactions, are performed with a PHP backend as part of a LAPP (Linux, Apache, PostgreSQL, PHP) stack architecture. This system is built using a Model-View-Controller design to appropriately relate processes together. A PostgreSQL relational database is incorporated to store user metadata and overall time series dataset information. CILogon is used for identity management and security of user credentials when accessing the site[22].

For other, more technical processes, such as analytical processing, job creation, and LLM interactions, a separate Python Flask service was developed and incorporated as well. This separation allows the platform to leverage robust Python libraries and API interactions for more computationally expensive or complex tasks. The PHP web service backend interacts with this Python server using the Guzzle HTTP client to make requests to the Flask server's RESTful API. The LLM is queried using an OpenAI-compatible API, meaning any quality reasoning model can be utilized. In the Forecaster's current implementation, the DeepSeek-R1 model is used[23], implemented and hosted locally at the University of Kentucky to ensure that all data remains secure and private. This implementation of the model leverages an 8-GPU H200 cluster with 141GB VRAM per GPU to host the models' weights. DeepSeek was chosen due to its open source nature and ability to perform complex reasoning using external context, but any commercial or open source reasoning models could be substituted.

The training and evaluation of the chosen models is also performed with a separate Python agent. This agent is independent of the machine that the web services are running on and can be set up on any other machine in the network. The connection between this agent and the website is created using ClearML[24], an infrastructure platform that can manage training jobs by collecting parameters and sending information to agents in the system. The Python agent can also send API requests back to the PHP backend to give updates on the training process, which are made visible to the user through the web interface.

Object storage for data and results is implemented with a self-hosted, S3-compatible system. This stores all large data artifacts, including raw user-uploaded time series files and pickled trained models. Both backends can interact with this data store, using the AWS S3Client library in PHP[25] and the boto3 library in Python[26]. Thus, these files can be accessed both by the web service and by the separate job agent. This overall system is represented in Figure 1.

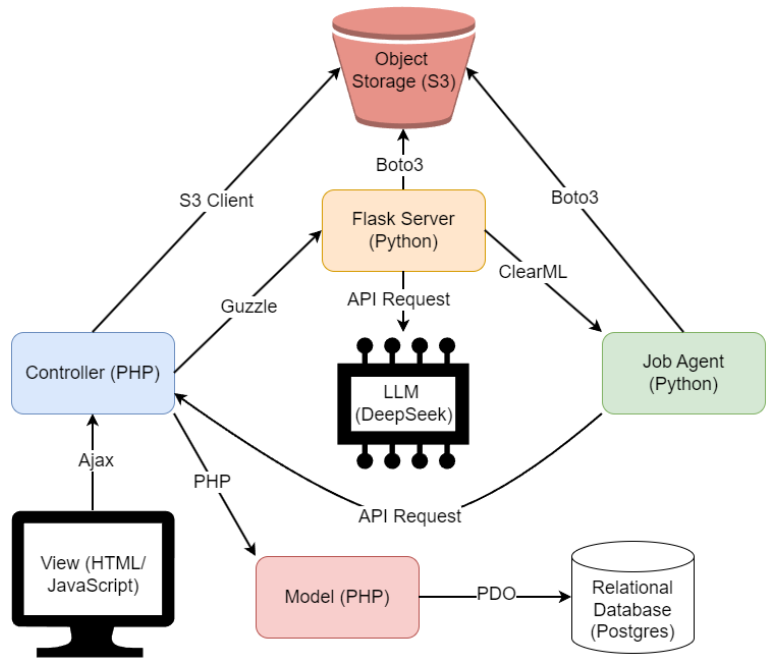

**Figure 1.** Architecture of the Forecaster system.

To begin the process of running a job, a user must first upload a .csv file containing their data. When the user uploads this file, the system validates file size and naming requirements. If the file is valid, a call is made to the controller to save the file to the object storage system, using a combination of the user ID and filename to create a unique identifier (users cannot upload more than one file of the same name). A record is also made in the PostgreSQL database that links the dataset with the user and contains other metadata, such as the time of upload. Additionally, the column names are extracted from the .csv and returned to the view to be displayed on the next page.

The user can then classify each of the columns of their dataset according to their roles in the forecasting task. The possible roles are 'Not Included', 'Time Component', 'Grouping', 'Target', 'Past Covariate', 'Future Covariate', and 'Static Covariate'. 'Not Included' indicates a column that should be ignored. The 'Time Component' column represents the date or timestep for the time series. 'Grouping' indicates a column that distinguishes between multiple groupings that should be simultaneously forecasted; for example, a dataset may contain flu cases for different countries, and the user wants to create forecasts for each country. In that case, the column indicating the country would be the 'Grouping' column. The 'Target' can be one or multiple columns that should be forecasted. If multiple columns are selected, these components will be analyzed and forecasted using a single multivariate model, except for some models that will be described later that can only perform univariate forecasting, in which case multiple models will be trained. The covariate columns indicate variables that should be used by the models to help with predictions by providing additional trends to analyze and compare. 'Past Covariates' are variables for which only previous values would be known, but values at prediction time would be unknown. Variables known ahead of prediction time, like day of the week or planned promotions, are known as 'Future Covariates'. 'Static Covariates' are variables that stay constant but can help distinguish between groupings.

To assist with the user's choices during this stage, variables can be plotted using an additional tool provided on this page of the website. The user can select a variable (typically the time component) for the x-axis, and one or more variables to plot along the y-axis. With this functionality, users can analyze the columns and compare features to determine how to best categorize those columns. Typically, these plots will take the form of a line graph, unless a categorical variable is chosen. If so, a Gantt-like chart will be produced instead. The plotting is performed using the JavaScript Plotly library[27]. Once the data categories are submitted, a JSON string containing each column name and its corresponding category is saved to the PostgreSQL database. Figure 2 illustrates the page design, featuring an example using a publicly available dataset containing daily counts of bike riders in Norway[28].

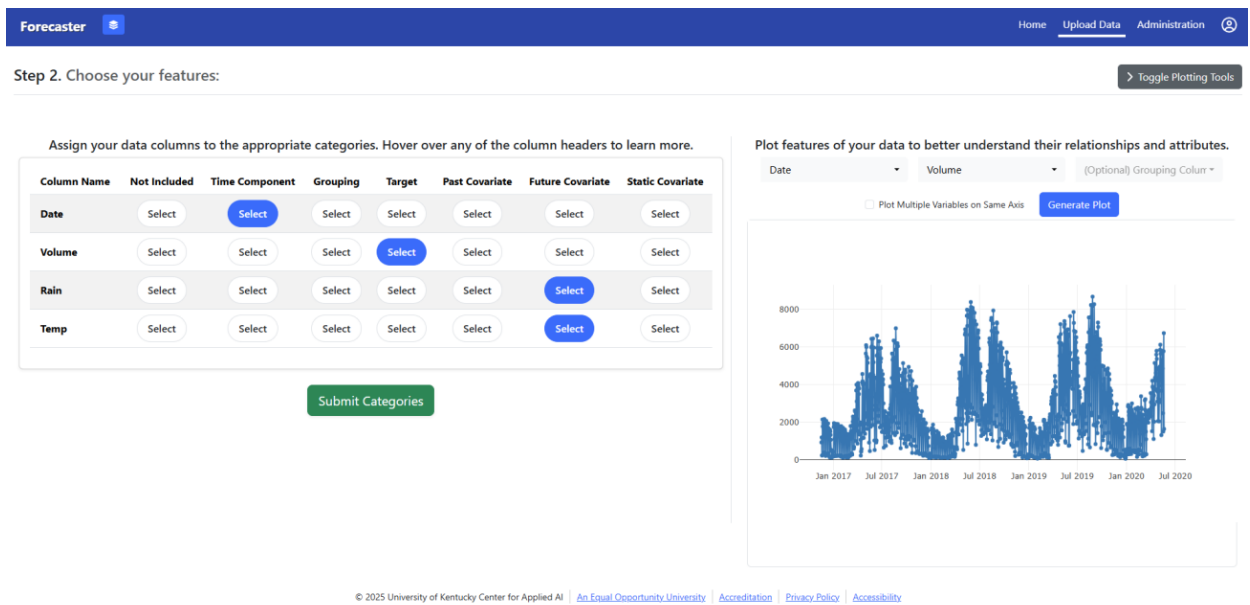

**Figure 2.** Displays columns and plotted data, with the plotting tab able to be shown or hidden on the page.

After submitting these categories, the next page allows the user to make choices regarding the training job and parameters. The models the user can select include ARIMA[29], exponential smoothing[30], linear regression[31], XGBoost[32], random forest[33], light gradient boosting machine[34], N-Linear model[35], and temporal fusion transformer[36]. Some of these models have different limitations; for example, the ARIMA model cannot use future covariates, the temporal fusion transformer must have future covariates, and this implementation of exponential smoothing cannot use any covariates and can only train on one time series at a time. These models have different parameters that can be modified. Only the relevant parameters to the models that have been selected for training will be shown. Some parameters apply to many models, such as choices for input/output chunk length or number of epochs for training.

The user can also make other choices on this page that affect the training and evaluation process. Users can choose if the models should, when possible, generate probabilistic forecasts, meaning that a distribution function is chosen, and many predictions are made for each timestep before being averaged together. The user can also choose how the models will be trained and evaluated. One option will evaluate the model over a holdout test set at the end of the series. Another option, if the user is confident with the model and data, is to use the entire series for training and generate predictions for a certain number of timesteps after the end of the timeseries. Users can also choose to do this on a later page once their models have been fully trained. Finally, the user can choose an expanding window method of training and evaluation. With this technique, each model will be trained only on the beginning of the timeseries and will produce forecasts up to a certain horizon after this subset. Subsequently, the training window will expand a certain number of timesteps and retrain, learning from the errors in the previous forecast, and it will generate new predictions over the next period. This cycle continues until the model has trained on the entire dataset.

Choosing the correct parameters for training can be difficult because the 'correct' choices depend on the size and structure of the dataset. The period of the timeseries can have a large influence on proper parameter choices, as specific values should be guided by the natural cycles in the timeseries data. For example, it would make sense to consider hourly data in 12 or 24 timestep windows because this aligns with natural daily cycles. If the timeseries has a daily period, then increments of 7 or 30 align with weekly or monthly periods, which frequently contribute important seasonality patterns to the data. Additionally, the size of the data and presence of certain types of covariates can change what models would be recommended to work with. Thus, it was important to incorporate a process for obtaining recommendations for parameter and model choices.

This was accomplished using an LLM. If the user selects 'Get Recommendations', a request will be made to a Flask server function that reads in the data and produces overall summary statistics. This includes the number of observations, the frequency of the time component, and statistics about the target variable, such as mean, distribution, and proportion of zeros. These statistics are then passed as context to a request made to the DeepSeek-R1 model. The LLM is instructed in the prompt to provide recommendations in a JSON format for a given list of parameters and models. Thus, the output from the LLM call contains keys for each parameter, which can be used to automatically fill in the values of the appropriate input elements on the website. The user is free to further modify these parameter choices. In addition, the LLM is instructed to write a brief explanation for the parameter picks, which is provided to the user through a text box on the webpage. An example screenshot for this webpage is provided in Figure 3.

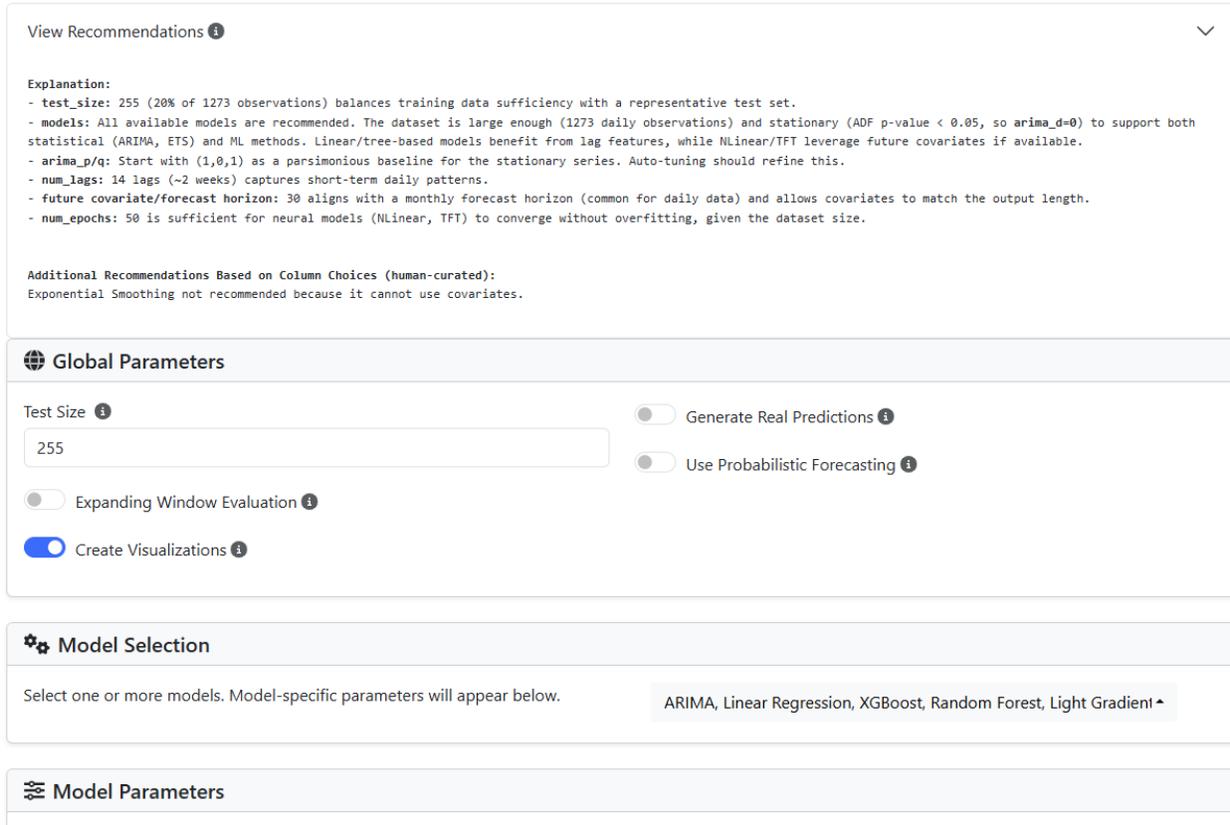

**Figure 3.** Section of parameter page with LLM recommendations.

Once satisfied with the choices, the user can begin the training job. When this happens, a request is first made to the PHP controller that contains the parameter choices from the webpage; the controller will also collect the dataset column categorizations from the PostgreSQL database. All of the choices are sent as arguments to a ClearML job. This includes information about the dataset necessary for the agent to pull the correct timeseries data from the object storage. When this ClearML job begins, it will enter a queue. Any running agent that is not actively performing a job will be listening to this queue. Thus, multiple agents can be set up to account for higher demand in the system.

When an agent begins a job, it pulls down the parameters from ClearML and uses the information to pull the dataset from the object storage. It uses the column distinctions to classify each column of the data appropriately. The Python agent primarily uses the Darts library to build timeseries objects and train and evaluate each model type[37]. Once the dataset is read in and formatted correctly, the target variable(s) and covariates, if applicable, are individually normalized to range from 0 to 1. Then, each model is iterated through, producing predictions for the specified time frame. Additionally, a Naïve Seasonal model that only copies the most recent seasonal values of the timeseries for its prediction is trained. This provides a baseline of results for comparison with the other models.

After each model is trained and produces predictions, each prediction set is compared to the true values. Performance metrics are calculated over the test set, or if the expanding window method was chosen, they are calculated over all timesteps for which the model was evaluated during the training process. Two versions of each performance metric are calculated: a normalized value calculated over the scaled series and a de-normalized value calculated over the original, un-modified series. The normalized metrics are helpful for comparing performance between groupings and components because they are independent of the scale of the data. The de-normalized metrics follow the scale of the data and are more intuitive for real interpretations. Both the predicted and real target values for each model, as well as the performance metrics, are saved in a JSON format to the object storage.

Throughout the training process, the agent sends API requests to the frontend PHP controller, modifying the view for that dataset on the home page of the website to provide updates on how many models have been trained. This allows the user to follow the progress of their job. Once the job has completed, a final request is sent, and the frontend is notified to update the home page to allow the user to view results for that job.

On the results page, an overall table shows the performance metrics for each model. The columns for each metric are sortable and toggleable so the user can view only the metrics they are interested in. If the dataset contains multiple groupings or components, dropdowns are provided to allow the user to look at an overall summary or view metrics specific to each grouping/component. An output log is also viewable on this page and allows the user to see any warnings or errors that occurred during the job. The user can generate visualizations for each model, which will show the training data, testing data (if applicable), and predictions in a Plotly-generated graph. These graphs are generated at runtime using the JSON data stored in the object storage. Users can zoom and toggle different components and groupings to get more detailed views of segments of data. These graphs can also be easily exported as images.

Interpreting the performance metrics for each model can be overwhelming, so LLM assistance is also available on this page. If the user clicks the 'Summarize Results' button, a JSON of the performance metric data is passed to the DeepSeek-R1 model, and the model is instructed to summarize these results in an easy-to-understand way and provide recommendations and explanations for which models appear best suited to the data. The LLM's response is then shown in a text box on the results page, similarly to the parameter recommendations.

New predictions can be generated on this page using the trained models. The user can select which models to use and how many timesteps to forecast for. A new .csv file can be uploaded containing future covariate values, and the models will use these future values to make predictions for the new timesteps. This process operates similarly to the training workflow. A job is created in ClearML for the prediction task, and parameters are passed through ClearML to a listening agent. The agent reads in the dataset, trained model(s), and any future covariates from the object storage. If the models were initially trained on future covariate data but this data is not provided or does not cover the entire forecast horizon, then it will be imputed for the modern timesteps using the Pandas interpolate method, which fills in new data assuming a linear trend[38]. Once the covariate data is handled, the models will be iterated through to produce new predictions for the specified number of timesteps, and these predictions are saved to object storage.

Once the forecasting job is completed, the user can download the results as a JSON file, which contains raw forecast counts for each model. Additionally, if the user generates a visualization on the interface, it will now include the new forecast data as part of the resulting line graph for each applicable model.

**Results**

The Forecaster was tested across multiple kinds of datasets to ensure performance was sufficient across different data types and complexities. The quality of the specific results on these datasets is not the focus of this paper, but the results are included to demonstrate the kinds of metrics available to users and to justify the effectiveness of the system.

Four different datasets were used to test the forecasting performance of the site. The first, already described through the examples in the previous section, contains daily counts of bike riders in Norway, with temperature and precipitation columns used as covariates. The second is a simple, one-dimensional time series of electrocardiogram (ECG) readings sampled at a rate of 100 Hz[39]. The third dataset contains daily sales information for stores in Ecuador, containing counts of sales for different categories of products, as well as a covariate indicating whether that category of product had an active promotion for that day[40]. This dataset was limited to only a single store for this analysis, and the categories of products were considered distinct groupings. The fourth dataset contains daily COVID-19 data for a variety of countries. This dataset contains multiple series that can be forecasted, such as new cases and new deaths resulting from COVID-19, and a variety of covariates, such as new vaccination rates and static covariates for each country like population density, median age, and rates of cardiovascular disease and smoking[41].

These datasets were chosen to ensure a variety of industries, use cases, and dataset types were tested. It creates a comprehensive evaluation of different types of covariates (past, future, static) and ensures multiple temporal periods, grouping types, and target variable distributions are included. Table 1 shows the normalized results for the best-performing model for each dataset, using both kinds of evaluation (train/test split, expanding window). The best-performing model was chosen as the one with the lowest values across the most different metrics. For the expanding window evaluation, the metrics were calculated across the entire series, excluding the initial segment used for training only. For datasets with multiple groupings (store sales, COVID-19 cases) or multiple target series (new COVID-19 cases and new deaths), the metrics are averaged across these distinctions.

| Dataset | Evaluation Type | Mean Absolute Error (MAE) | Mean Absolute Percentage Error (MAPE) | Mean Squared Error (MSE) | Root Mean Squared Error (RMSE) | Symmetric Mean Absolute Percentage Error (SMAPE) | Mean Absolute Scaled Error (MASE) | Best Performing Model |
|---|---|---|---|---|---|---|---|---|
| **Bike rides** | Train/test | 0.110 | 0.000* | 0.021 | 0.143 | 0.000* | 0.944 | Random Forest |
| | Expanding Window | 0.086 | 0.000* | 0.014 | 0.117 | 0.000* | 1.184 | Random Forest |
| **ECG** | Train/test | 0.031 | 7.845 | 0.007 | 0.085 | 8.095 | 24.235 | Random Forest |
| | Expanding Window | 0.064 | 28.518 | 0.013 | 0.112 | 19.065 | 16.152 | ARIMA |
| **Sales** | Train/test | 0.071 | 0.000* | 0.011 | 0.099 | 0.000* | 0.634 | Exponential Smoothing |
| | Expanding Window | 0.065 | 0.000* | 0.009 | 0.094 | 0.000* | 0.766 | Exponential Smoothing |
| **COVID-19** | Train/test | 0.393 | 215.887 | 0.923 | 0.615 | 77.589 | 50.386 | ARIMA |
| | Expanding Window | 0.224 | 70.045 | 0.313 | 0.394 | 45.724 | 1.000* | XGBoost |

**Table 1.** Summary of results across datasets and evaluation types.

Table 1 shows that the expanding window method typically leads to better performance during evaluation than the standard train/test split technique. In some cases, the MAPE and SMAPE values equal zero because the series they are calculated for contains values of zero, which causes the calculation for these percentage metrics to return an undefined value. Additionally, the MASE fails in rare cases due to discrepancies between the training set length and the length of the seasonality period of the time series. Each dataset also has inherent differences that influence the performance of each model. The COVID-19 dataset performs the worst overall when averaged together, but this is because several of the countries have unprecedented increases in cases towards the end of the series, which inflates the error metrics during those periods. The country with the most accurate predictions was India, where the XGBoost model achieved metrics much closer to the results of the other datasets, such as an RMSE of 0.106.

Examples of visualizations showcasing the predictions are given in Figure 5. Figure 5(A) shows the performance of the random forest model when evaluated just on the test set for the bike riding dataset, while Figure 5(B) shows the expanding window evaluation for the same model.

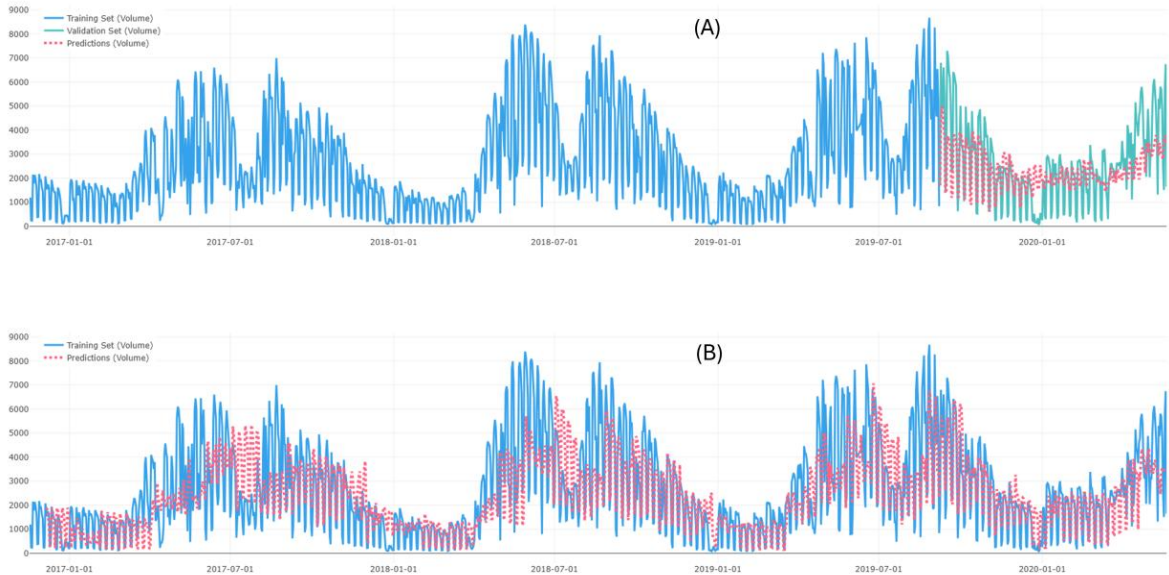

**Figure 5.** Visualizations of predictions on bike riding dataset, for both train/test (A) and expanding window (B).

**Discussion**

The Forecaster system's architecture allows for applicability to a variety of datasets like the ones tested here for demonstration. Results can be easily compared between datasets and models due to the variety of metrics produced in both normalized and de-normalized ways. Also, jobs can be easily re-run with different covariates or parameters to compare performance on the same dataset.

Across the tested datasets, the same models would frequently be among the best or worst performing. Models such as random forest, XGBoost, and ARIMA were typically the ones that created the most accurate predictions. This likely speaks more to their applicability to a variety of datasets and complexities than it does to any fundamental truth about forecasting models in general. Neural network models like N-Linear LSTF and Temporal Fusion Transformer frequently performed among the worst models here, but there could be several reasons for this. These models may be more complex than the tested datasets require, and they would be better suited to much larger time series. All of the models tested here were run with default parameters. These complex neural network models are more dependent on appropriate parameter choices, meaning they may underperform when time is not spent tuning the necessary arguments. The Forecaster interface allows for modification of different parameters, so tuning these through the website is likely a necessary step to achieve optimal performance with these kinds of models.

In the future, more work will be done to expand the flexibility of the input data format, rather than limiting the uploads to only .csv files. There are plans to develop an intake system that allows for continually updating streams of data to be read in from timeseries databases. This would support the ability to make repeated predictions on a time series that continually extends with new data, ensuring that the predictions generated stay updated and relevant. An API system is planned to be further developed that would allow for programmatic interactions with the system, supporting the incorporation of this tool with learning health systems and continually streaming data sources.

The Forecaster has limitations concerning full model customization or advanced feature engineering techniques. These complex additions would come at the expense of usability for non-technical researchers, and the tool was designed with the primary goal of making forecasting accessible. Experienced users hoping to implement more complex techniques can still utilize the Forecaster for exploratory analysis, plotting, and generation of model and data files that can be downloaded and further explored with custom code.

## Conclusion

The Forecaster is a powerful tool for training and evaluating time series forecasting models, as well as interpreting data and results through visualizations, wrapped into an easy-to-use web interface. It provides no-code solutions to researchers and clinicians who wish to take advantage of the available tools for forecasting without possessing the required technical background to implement complex models. This interface allows users to simply upload data, choose data options and parameters, and begin a training job without needing to worry about required computational resources or writing code. Parameters and results can be explained on the website with an LLM, lowering the barrier to understanding how these models work or what their results mean. This system is planned to be incorporated within larger frameworks for continuous health monitoring using data streams and repeated re-training and evaluation of forecasting models. The Forecaster has a variety of applications, especially in healthcare research, where time series data is common and the ability to create accurate forecasts can have strong positive effects on health outcomes.

## Acknowledgement

This research was supported in part by the National Institutes of Health under award number UL1TR001998. The content is solely the responsibility of the authors and does not necessarily represent the official views of the NIH.

## References

1. Praggnya Kanungo. Time series forecasting in financial markets using Deep Learning Models. World Journal of Advanced Engineering Technology and Sciences. 2025 Apr 30;15(1):709–19. doi:10.30574/wjaets.2025.15.1.0167
2. Sun G, Deng S. Financial time series forecasting: A comparison between traditional methods and AI-driven techniques. Journal of Computer, Signal, and System Research. 2025 Mar 28;2(2):86–93. doi:10.71222/339b9812
3. Sokannit P. Forecasting household electricity consumption using time series models. International Journal of Machine Learning and Computing. 2021 Nov;11(6):380–6. doi:10.18178/ijmlc.2021.11.6.1065
4. Neumann O, Turowski M, Mikut R, Hagenmeyer V, Ludwig N. Using weather data in energy time series forecasting: The benefit of input data transformations. Energy Informatics. 2023 Nov 2;6(1). doi:10.1186/s42162-023-00299-8
5. Willard JD, Varadharajan C, Jia X, Kumar V. Time series predictions in unmonitored sites: A survey of machine learning techniques in water resources. Environmental Data Science. 2025;4. doi:10.1017/eds.2024.1
6. Mansur S, Sattar K, Hosseini SE, Pervez S, Ahmad I, Saleem K, et al. Sales forecasting for retail stores using hybrid neural networks and sales-affecting variables. PeerJ Computer Science. 2025 Sept 11;11. doi:10.7717/peerj-cs.3058
7. Tomov L, Chervenkov L, Miteva DG, Batselova H, Velikova T. Applications of time series analysis in Epidemiology: Literature Review and our experience during COVID-19 pandemic. World Journal of Clinical Cases. 2023 Oct 16;11(29):6974–83. doi:10.12998/wjcc.v11.i29.6974
8. Patharkar A, Cai F, Al-Hindawi F, Wu T. Predictive modeling of Biomedical Temporal Data in healthcare applications: Review and Future Directions. Frontiers in Physiology. 2024 Oct 15;15. doi:10.3389/fphys.2024.1386760
9. Hochreiter S, Schmidhuber J. Long Short-Term Memory. Neural Computation. 1997 Nov 1;9(8). doi:https://doi.org/10.1162/neco.1997.9.8.1735
10. Rudin C. Stop explaining black box machine learning models for high stakes decisions and use interpretable models instead. Nature Machine Intelligence. 2019 May 13;1(5):206–15. doi:10.1038/s42256-019-0048-x
11. Ahmad M, Rehman AA, Khan R, Bibi H. Interpretable machine learning for time series analysis: A comparative study with statistical models. ACADEMIA International Journal for Social Sciences. 2025 Aug 28;4(3):4001–9. doi:10.63056/acad.004.03.0681
12. Furizal F, Ma'arif A, Kariyamin, Firdaus AA, Wijaya SA, Nakib AM, et al. Understanding Time Series Forecasting: A Fundamental Study. Buletin Ilmiah Sarjana Teknik Elektro. 2025 Oct;7. doi:10.12928/biste.v7i3.13318
13. Introduction [Internet]. [cited 2025 Oct 29]. Available from: https://www.statsmodels.org/stable/index.html
14. Learn [Internet]. [cited 2025 Oct 29]. Available from: https://scikit-learn.org/stable/
15. Pytorch [Internet]. [cited 2025 Oct 29]. Available from: https://pytorch.org/

16. Home [Internet]. 2023 [cited 2025 Oct 29]. Available from: https://pycaret.org/
17. [Internet]. [cited 2025 Oct 29]. Available from: https://aws.amazon.com/forecast/
18. Time-series modeling [Internet]. [cited 2025 Oct 29]. Available from: https://docs.datarobot.com/en/docs/modeling/time/index.html
19. Train a model with tabular workflow for forecasting | vertex AI | google cloud [Internet]. Google; [cited 2025 Oct 29]. Available from: https://cloud.google.com/vertex-ai/docs/tabular-data/tabular-workflows/forecasting-train
20. Vishwas BV, Macharla SR. Time-LLM: Reprogramming large language model. Time Series Forecasting Using Generative AI. 2025;131–54. doi:10.1007/979-8-8688-1276-7_4
21. Tang H, Zhang C, Jin M, Yu Q, Wang Z, Jin X, et al. Time series forecasting with LLMS: Understanding and enhancing model capabilities. ACM SIGKDD Explorations Newsletter. 2025 Jan 21;26(2):109–18. doi:10.1145/3715073.3715083
22. [Internet]. [cited 2025 Oct 30]. Available from: https://www.cilogon.org/
23. DeepSeek-AI, Guo D, Yang D, Zhang H, Song J, Zhang R, et al. DeepSeek-R1: Incentivizing Reasoning Capability in LLMs via Reinforcement Learning. 2025 Jan 22; doi:https://doi.org/10.48550/arXiv.2501.12948
24. AI Infrastructure Platform: Maximize AI Performance & Scalability [Internet]. 2025 [cited 2025 Oct 30]. Available from: https://clear.ml/
25. [Internet]. [cited 2025 Oct 30]. Available from: https://docs.aws.amazon.com/aws-sdk-php/v3/api/class-Aws.S3.S3Client.html
26. Boto3 documentation¶ [Internet]. [cited 2025 Oct 30]. Available from: https://boto3.amazonaws.com/v1/documentation/api/latest/index.html
27. Data apps for production [Internet]. [cited 2025 Oct 30]. Available from: https://plotly.com/
28. Banachewicz K. Norway bicycles [Internet]. 2021 [cited 2025 Oct 30]. Available from: https://www.kaggle.com/datasets/konradb/norway-bicycles/data
29. Arima [Internet]. [cited 2025 Oct 30]. Available from: https://unit8co.github.io/darts/generated_api/darts.models.forecasting.arima.html
30. Statsmodels.tsa.holtwinters.ExponentialSmoothing [Internet]. [cited 2025 Oct 30]. Available from: https://www.statsmodels.org/stable/generated/statsmodels.tsa.holtwinters.ExponentialSmoothing.html
31. Linear Regression Model¶ [Internet]. [cited 2025 Oct 30]. Available from: https://unit8co.github.io/darts/generated_api/darts.models.forecasting.linear_regression_model.html
32. Chen T, Guestrin C. XGBoost. Proceedings of the 22nd ACM SIGKDD International Conference on Knowledge Discovery and Data Mining. 2016 Aug 13;785–94. doi:10.1145/2939672.2939785
33. Randomforestregressor [Internet]. [cited 2025 Oct 30]. Available from: https://scikit-learn.org/stable/modules/generated/sklearn.ensemble.RandomForestRegressor.html#sklearn.ensemble.RandomForestRegressor
34. Ke G, Meng Q, Finley T, Wang T, Chen W, Ma W, et al. LightGBM: a highly efficient gradient boosting decision tree. Proceedings of the 31st International Conference on Neural Information Processing Systems. 2017;3149–57.
35. Zeng A, Chen M, Zhang L, Xu Q. Are transformers effective for time series forecasting? Proceedings of the AAAI Conference on Artificial Intelligence. 2023 Jun 26;37(9):11121–8. doi:10.1609/aaai.v37i9.26317
36. Lim B, Arık S, Loeff N, Pfister T. Temporal Fusion Transformers for interpretable multi-horizon time series forecasting. International Journal of Forecasting. 2021 Oct;37(4):1748–64. doi:10.1016/j.ijforecast.2021.03.012
37. Time series made easy in python [Internet]. [cited 2025 Oct 30]. Available from: https://unit8co.github.io/darts/
38. Pandas [Internet]. [cited 2025 Oct 30]. Available from: https://pandas.pydata.org/
39. Andrikov D. ECG timeseries for prediction [Internet]. 2025 [cited 2025 Nov 6]. Available from: https://www.kaggle.com/datasets/denisandrikov/ecg-timeseries-for-prediction
40. Store sales - time series forecasting [Internet]. [cited 2025 Nov 6]. Available from: https://www.kaggle.com/competitions/store-sales-time-series-forecasting/data?select=train.csv
41. Anttal TS. Covid 19 dataset till 22/2/2022 [Internet]. 2022 [cited 2025 Nov 6]. Available from: https://www.kaggle.com/datasets/taranvee/covid-19-dataset-till-2222022